\documentclass[acmsmall, nonacm, screen]{acmart}

\title{Enhancing Pancreatic Cancer Staging with Large Language Models: The Role of Retrieval-Augmented Generation}

\makeatletter
\g@addto@macro\@authornotes{\footnotetext{(1) Department of Radiology, University of Yamanashi, Yamanashi, Japan.}}
\g@addto@macro\@authornotes{\footnotetext{(2) Department of Gastroenterology and Hepatology, University of Yamanashi, Yamanashi, Japan.}}
\g@addto@macro\@authornotes{\footnotetext{(3) Department of Radiation Oncology, Tohoku University Graduate School of Medicine, Sendai, Japan.}}
\g@addto@macro\@authornotes{\footnotetext{(4) Center for Medical Education and Sciences, University of Yamanashi, Yamanashi, Japan.}}
\makeatother

\author{Hisashi Johno$^\text{(1)}$}
\authornote{
  Corresponding author: Hisashi Johno, M.D., Ph.D.
  Department of Radiology, University of Yamanashi.
  1110 Shimokato, Chuo, Yamanashi, 409-3898, Japan.
  E-mail: johnoh@yamanashi.ac.jp
}
\author{Yuki Johno$^\text{(2)}$}
\author{Akitomo Amakawa$^\text{(1)}$}
\author{Junichi Sato$^\text{(1)}$}
\author{Ryota Tozuka$^\text{(1)(3)}$}
\author{Atsushi Komaba$^\text{(1)}$}
\author{Hiroaki Watanabe$^\text{(1)}$}
\author{Hiroki Watanabe$^\text{(1)}$}
\author{Chihiro Goto$^\text{(1)}$}
\author{Hiroyuki Morisaka$^\text{(1)}$}
\author{Hiroshi Onishi$^\text{(1)}$}
\author{Kazunori Nakamoto$^\text{(4)}$}
\authorsaddresses{}

\usepackage[capitalize]{cleveref}
\usepackage[
  list-final-separator={, \text{and} }
]{siunitx}
\usepackage{tabularx}
\usepackage{tikz}
\usetikzlibrary{positioning}
\usepackage{enumitem}
\setlist{nosep, leftmargin=1em}
\setlist[itemize, 1]{label={\tikz[baseline=-2pt] \fill circle[radius=1pt];}}
\setlist[itemize, 2]{label={\tikz[baseline=-2pt] \draw circle[radius=1pt];}}
\setlist[itemize, 3]{label={\tikz[baseline=-2pt] \fill (1pt, 1pt) rectangle (-1pt, -1pt);}}
\usepackage{soul}
\setulcolor{red}

\begin{document}
  \maketitle
  \renewcommand{\shortauthors}{Johno et al.}

  \section*{Abstract}
    \subsection*{Purpose}
      Retrieval-augmented generation (RAG) is an emerging technology to enhance the functionality and reliability of large language models (LLMs) by retrieving relevant information from reliable external knowledge (REK). RAG has gained increasing interest in radiology, and we previously reported the utility of NotebookLM, an LLM with RAG (RAG-LLM), for lung cancer staging. However, since the comparator LLM differed from NotebookLM's internal model, it remained unclear whether NotebookLM's advantage stemmed from RAG techniques or inherent differences in LLM performance. To better isolate the impact of RAG and to assess its utility across different cancers, this study compared the performance of NotebookLM and its internal LLM, Gemini 2.0 Flash, in a pancreatic cancer staging experiment.

    \subsection*{Materials and methods}
      A paper summarizing Japan's current pancreatic cancer staging guidelines was used as REK. We compared the performance of three groups---REK+/RAG+ (NotebookLM with REK), REK+/RAG- (Gemini 2.0 Flash with REK), and REK-/RAG- (Gemini 2.0 Flash without REK)---in staging 100 fictional pancreatic cancer cases based on CT findings. Staging criteria included TNM classification, local invasion factors, and resectability classification. In the REK+/RAG+ group, retrieval accuracy was quantified based on the sufficiency of retrieved REK excerpts.

    \subsection*{Results}
      REK+/RAG+ achieved a staging accuracy of \qty{70}{\percent}, outperforming REK+/RAG- (\qty{38}{\percent}) and REK-/RAG- (\qty{35}{\percent}). For TNM classification, REK+/RAG+ attained \qty{80}{\percent} accuracy, exceeding that of REK+/RAG- (\qty{55}{\percent}) and REK-/RAG- (\qty{50}{\percent}). Additionally, REK+/RAG+ explicitly presented retrieved REK excerpts as the basis for its classifications, achieving a retrieval accuracy of \qty{92}{\percent}.

    \subsection*{Conclusion}
      NotebookLM, a RAG-LLM, outperformed its internal LLM, Gemini 2.0 Flash, in a pancreatic cancer staging experiment, suggesting that RAG may improve the staging accuracy of LLMs. Furthermore, NotebookLM effectively retrieved relevant REK excerpts, providing transparency for radiologists to verify response reliability and highlighting the potential of RAG-LLMs in supporting clinical diagnosis and classification.

    \section*{Keywords}
      Large language model (LLM),
      Retrieval-augmented generation (RAG),
      Reliable external knowledge (REK),
      NotebookLM,
      Gemini 2.0 Flash,
      Pancreatic cancer staging

  \section{Introduction}
    \begin{figure}
      \includegraphics[width=\textwidth]{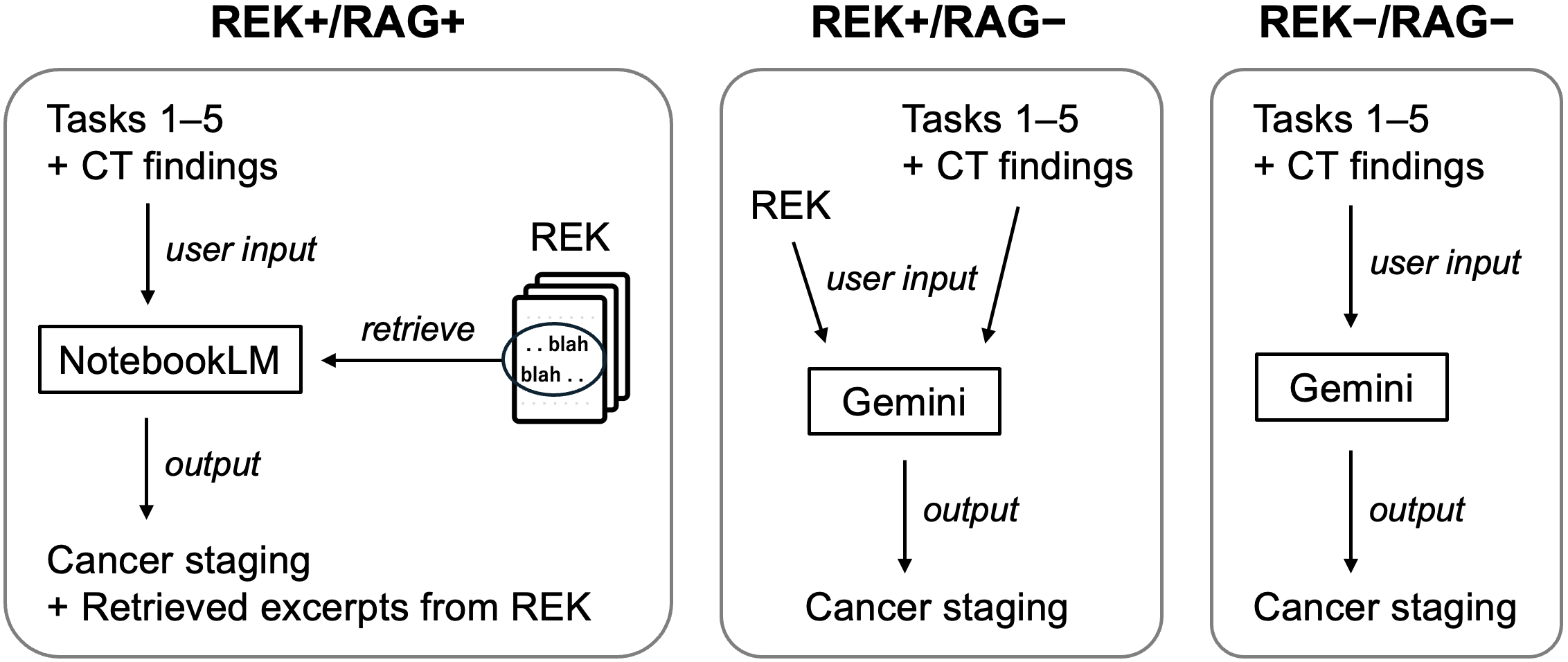}
      \caption{An overview of the experimental process. Radiologists from our team generated CT findings for 100 fictional pancreatic cancer patients. NotebookLM with REK (REK+/RAG+), Gemini 2.0 Flash with REK (REK+/RAG-), and Gemini 2.0 Flash without REK (REK-/RAG-) conducted cancer staging based on the CT findings in response to Tasks 1--5 (see Table 2). In the REK+/RAG+ group, retrieved excerpts from REK were available alongside the classifications. The REK was uploaded to the NotebookLM web system for RAG processing. In the REK+/RAG- group, the REK was manually entered into the prompt field before providing Tasks 1--5. \textit{REK}=reliable external knowledge, \textit{RAG}=retrieval-augmented generation}
      \label{fig1}
    \end{figure}

    Large language models (LLMs) have recently attracted attention in radiology, particularly for their potential to assist in image diagnosis and classification. However, their clinical application remains challenging, partly due to the risk of generating incorrect responses or providing answers unsupported by reliable evidence \cite{Keshavarz}. One strategy to address these challenges is retrieval-augmented generation (RAG), which enhances the accuracy and reliability of LLM-generated responses by retrieving relevant information from reliable external knowledge (REK) and incorporating it into the model's prompt \cite{Lewis,Shuster}. Research on LLMs with RAG (RAG-LLMs) in radiology is still in its early stages, and their effectiveness, particularly in image interpretation and classification, remains largely unexplored \cite{Bhayana2,Rau}.

    In a previous study, we evaluated the utility of NotebookLM (\url{https://notebooklm.google}), a RAG-LLM developed by Google, for lung cancer staging. We provided it with the latest Japanese lung cancer staging guidelines at the time as REK and tasked it with staging 100 fictional lung cancer cases based on CT findings. NotebookLM exhibited higher classification accuracy compared to GPT-4 Omni (GPT-4o). Furthermore, NotebookLM referenced the provided REK content with greater accuracy \cite{Tozuka}. These findings highlight NotebookLM's potential in cancer staging; however, its utility beyond lung cancer remains unclear. Moreover, a key limitation of this study was that the LLM used in NotebookLM at the time, Gemini 1.5 Pro, differed from the compared LLM, GPT-4o, meaning the comparison was not purely based on the presence or absence of RAG techniques.

    Therefore, in this study, we evaluated whether NotebookLM is also useful for staging pancreatic cancer as a different type of cancer. Additionally, by comparing it with Gemini 2.0 Flash (\url{https://gemini.google.com/app}), the LLM currently integrated into NotebookLM, we aimed to more purely assess the impact of RAG techniques while minimizing the influence of differences in the underlying language model.

  \section{Materials and methods}
    An overview of the experimental process is schematically summarized in \cref{fig1}. As shown in this figure, we compared the accuracy of cancer staging across three groups: REK+/RAG+ (NotebookLM with REK), REK+/RAG- (Gemini 2.0 Flash with REK), and REK-/RAG- (Gemini 2.0 Flash without REK). In the REK+/RAG+ group, retrieved excerpts from REK were available alongside the classification results. Therefore, we evaluated the specific content of these excerpts and assessed their relevance and appropriateness.

    \subsection*{Data preparation}
      \renewcommand{\thefootnote}{\fnsymbol{footnote}}
      Two radiologists from our team generated CT findings for 100 fictional pancreatic cancer patients, along with staging components (TNM classification, local invasion factors, and resectability classification) based on the latest pancreatic cancer staging guidelines in Japan---the eighth edition of the Japanese classification of pancreatic carcinoma \cite{Ishida}. The CT findings and staging components were subsequently reviewed and confirmed by four additional radiologists and one gastroenterologist. A breakdown of the staging components for the 100 fictional pancreatic cancer patients is provided in \cref{table1}. All the CT findings with staging components are available in Supplementary file 1\footnote[2]{Supplementary files 1 to 6 can be found in the ancillary files uploaded with this paper on arXiv.}. Below is an example from the dataset, presenting the first of the 100 cases:
      \begin{quote}
        Case 1 CT findings: {\ttfamily A nodular pancreatic cancer measuring 20 mm is observed in the body of the pancreas. No local invasion factors are noted. Lymph node metastases are identified in two nodes at station 10 and two nodes at station 11d. No other metastases are observed.}

        Case 1 staging components: T factor: T1c; N factor: N1b; M factor: M0; Local invasion factors: CH0, DU0, S0, RP0, PV0, A0, PL0, OO0; Resectability classification: R.
      \end{quote}
      \begin{table}
        \caption{Breakdown of staging components for the 100 fictional pancreatic cancer patients.}
        \label{table1}
        \begin{tabular}{l}
          \begin{tabular}{r*{8}{|c}}
            T factor & T0 & Tis & T1a & T1b & T1c & T2 & T3 & T4 \\
            \hline
            \phantom{Resectability classification}\llap{Number of patients} & 2 & 8 & 4 & 6 & 8 & 4 & 57 & 11
          \end{tabular} \\[1em]
          \begin{tabular}{r*{3}{|c}}
            N factor & N0 & N1a & N1b \\
            \hline
            \phantom{Resectability classification}\llap{Number of patients} & 58 & 22 & 20
          \end{tabular} \\[1em]
          \begin{tabular}{r*{2}{|c}}
            M factor & M0 & M1 \\
            \hline
            \phantom{Resectability classification}\llap{Number of patients} & 67 & 33
          \end{tabular} \\[1em]
          \begin{tabular}{r*{8}{|c}}
            Local invasion factors & CH1 & DU1 & S1 & RP1 & PV1 & A1 & PL1 & OO1 \\
            \hline
            \phantom{Resectability classification}\llap{Number of patients} & 7 & 4 & 62 & 64 & 47 & 34 & 7 & 16
          \end{tabular} \\[1em]
          \begin{tabular}{r*{3}{|c}}
            Resectability classification & R & BR & UR \\
            \hline
            Number of patients & 42 & 16 & 42
          \end{tabular}
        \end{tabular}
      \end{table}

    \subsection*{Preparation of REK and user input}
      We attempted to use an open-access paper \cite{Ishida}, which summarizes the current pancreatic cancer staging guidelines in Japan, as REK for NotebookLM and Gemini 2.0 Flash. However, since the full text of the paper (\num{5418} words, covering Chapters 1 to 8, including figure captions and tables) could not be entered into the prompt field at once, we adopted REK consisting of \num{4376} words from the paper (Chapters 1 to 6), omitting the final two chapters that were not relevant to staging. 

      To enable the LLM to perform pancreatic cancer staging, we provided Tasks 1--5, shown in \cref{table2}, as user input in the prompt field, followed by the CT findings for each case. In the REK+/RAG+ group, we uploaded the REK to the NotebookLM web system for RAG processing. In the REK+/RAG- group, we manually entered the REK into the prompt field before providing Tasks 1--5. In the REK-/RAG- group, we entered a prompt instructing adherence to the Japanese Classification of Pancreatic Carcinoma, Eighth Edition by the Japan Pancreas Society before providing Tasks 1--5.

      \begin{table}
        \caption{User input for LLMs to perform pancreatic cancer staging.}
        \label{table2}
        \begin{tabularx}{\textwidth}{l|X}
          Task 1 & Diagnose the local invasion factors of pancreatic cancer (CH, DU, S, RP, PV, A, PL, OO) and respond in the format: e.g., ``CH0, DU1, S1, RP1, PV0, A0, PL0, OO1''. \\
          \hline
          Task 2 & Based on the answer to Task 1, determine the T classification of pancreatic cancer (T0, Tis, T1a, T1b, T1c, T2, T3, T4). \\
          \hline
          Task 3 & Determine the N classification of pancreatic cancer (N0, N1a, N1b) based on the defined criteria for regional lymph nodes. Note that Metastasis to non-regional lymph nodes is classified under M classification, not N classification. \\
          \hline
          Task 4 & Determine the M classification of pancreatic cancer (M0, M1). \\
          \hline
          Task 5 & Based on the answer to Task 4, determine the resectability classification of pancreatic cancer (R, BR, UR). Respond with BR if the classification is BR-PV or BR-A, and UR if it is UR-LA or UR-M. If both BR and UR apply, respond with UR. \\
        \end{tabularx}
      \end{table}

    \subsection*{Evaluation}
      Cancer staging was defined as accurate if all staging components---TNM classification, local invasion factors, and resectability classification---were correctly determined. Staging accuracy was compared across the three groups: REK+/RAG+, REK+/RAG-, and REK-/RAG-. Additionally, the classification accuracy of each staging component was evaluated and compared across these groups. The TNM classification was considered correct only when all T, N, and M factors were accurately classified.

      In the REK+/RAG+ group, retrieved excerpts from REK via NotebookLM were available for reference. Therefore, we examined these excerpts for each case and evaluated retrieval accuracy. Retrieval was considered accurate if the excerpts contained sufficient information to correctly classify all of the staging components.

      The LLM's answers for each case in the three groups (REK+/RAG+, REK+/RAG-, and REK-/RAG-), along with the retrieved excerpts in the REK+/RAG+ group, are provided in Supplementary file 2\footnote[2]{\label{supplement:evaluation}Supplementary files 1 to 6 can be found in the ancillary files uploaded with this paper on arXiv.}. Additionally, the case-wise accuracy of each staging component across the three groups is presented in Supplementary file 3\footref{supplement:evaluation}, while the case-wise retrieval accuracy in the REK+/RAG+ group is provided in Supplementary file 4\footref{supplement:evaluation}.

    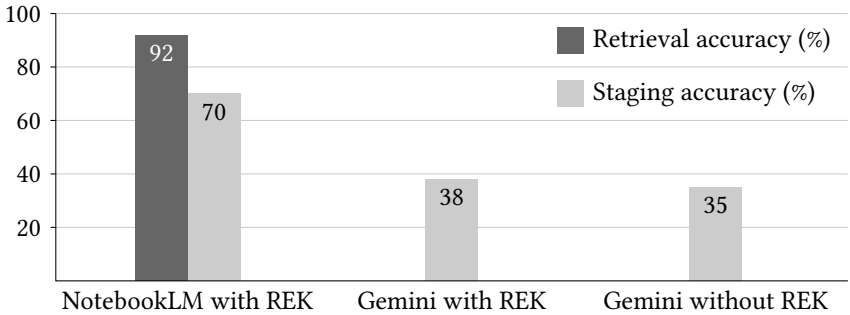
\begin{figure}
      \begin{tikzpicture}
        \foreach \y in {20, 40, 60, 80, 100}{
          \draw (0, 1pt * \y) -- +(-2pt, 0) node[left] {\y};
          \draw[black!20] (0, 1pt * \y) -- (300pt, 1pt * \y);
        }
        \node[below] at (50pt, 0) {NotebookLM with REK};
        \fill[black!20] (50pt, 0) rectangle (70pt, 70pt);
        \node[below] at (60pt, 70pt) {70};
        \fill[black!60] (30pt, 0) rectangle (50pt, 92pt);
        \node[below, white] at (40pt, 92pt) {92};
        \node[below] at (150pt, 0) {Gemini with REK};
        \fill[black!20] (140pt, 0) rectangle (160pt, 38pt);
        \node[below] at (150pt, 38pt) {38};
        \node[below] at (250pt, 0) {Gemini without REK};
        \fill[black!20] (240pt, 0) rectangle (260pt, 35pt);
        \node[below] at (250pt, 35pt) {35};
        \fill[black!60] (190pt, 85pt) rectangle (200pt, 95pt);
        \node[right] at (200pt, 90pt) {Retrieval accuracy (\unit{\percent})};
        \fill[black!20] (190pt, 65pt) rectangle (200pt, 75pt);
        \node[right] at (200pt, 70pt) {Staging accuracy (\unit{\percent})};
        \draw (0, 100pt) -- (0, 0) -- (300pt, 0);
      \end{tikzpicture}
      \caption{Staging performance of NotebookLM with REK, Gemini 2.0 Flash with REK, and Gemini 2.0 Flash without REK in the experiment using 100 fictional pancreatic cancer cases. Staging was considered accurate if all the staging components---TNM classification, local invasion factors, and resectability classification---were correctly determined. For NotebookLM, retrieval accuracy was also evaluated. Retrieval was considered accurate if the retrieved excerpts from REK contained sufficient information to enable accurate cancer staging. \textit{REK}=reliable external knowledge}
      \label{fig2}
    \end{figure}

  \section{Results}
\begin{figure}
      \includegraphics[width=\textwidth]{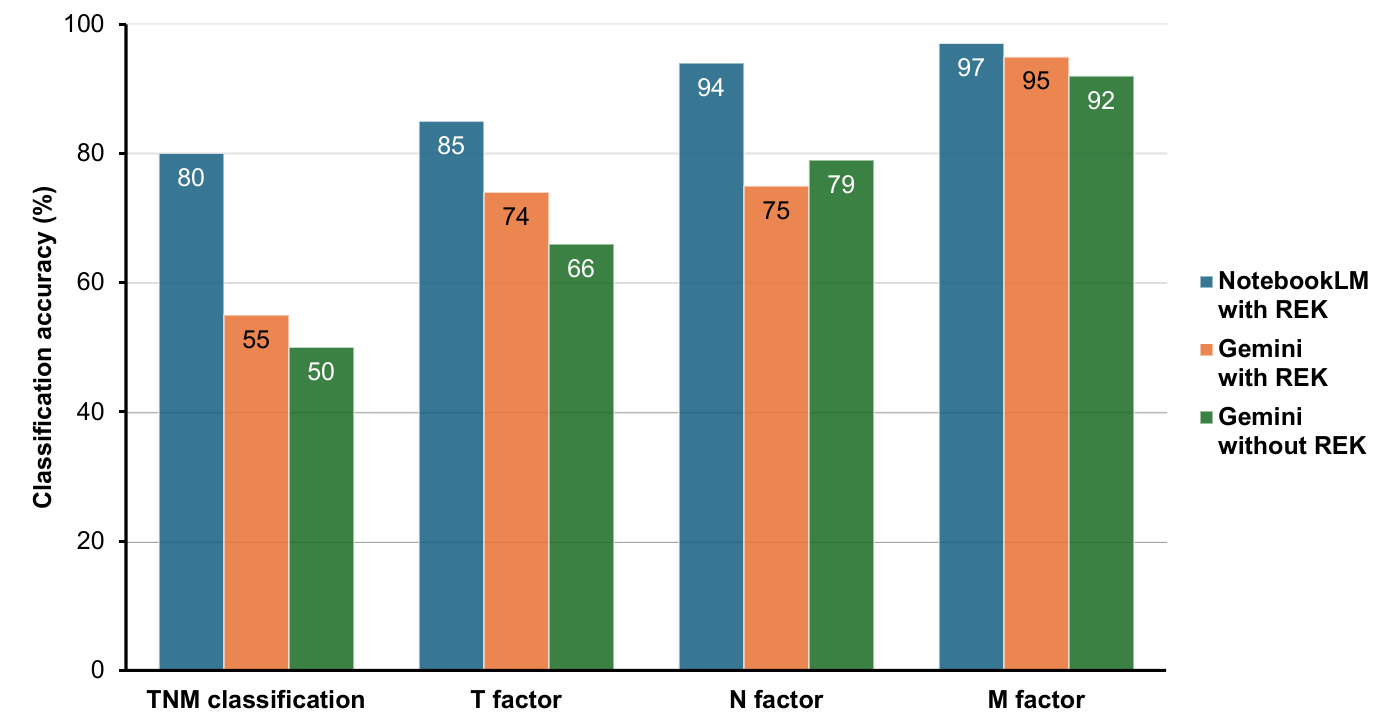}
      \caption{TNM classification performance of NotebookLM with REK, Gemini 2.0 Flash with REK, and Gemini 2.0 Flash without REK in the experiment using 100 fictional pancreatic cancer cases. The TNM classification was deemed correct only if all T, N, and M factors were accurately identified. Additionally, the classification accuracy for each T, N, and M factor was compared across the three groups. \textit{REK}=reliable external knowledge}
      \label{fig3}
    \end{figure}

\begin{figure}
      \includegraphics[height=22em]{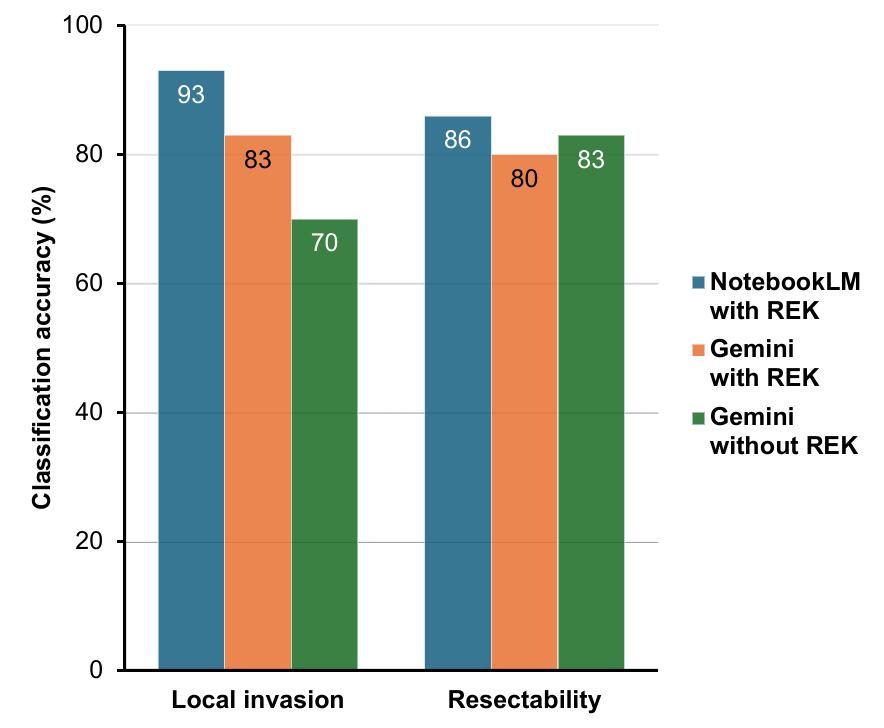}
      \caption{The performance of NotebookLM with REK, Gemini 2.0 Flash with REK, and Gemini 2.0 Flash without REK in determining local invasion factors and resectability classification in the experiment using 100 fictional pancreatic cancer cases. \textit{REK}=reliable external knowledge}
      \label{fig4}
    \end{figure}

    In the experiment using 100 fictional pancreatic cancer cases, NotebookLM with REK (REK+/RAG+) achieved a staging accuracy of \qty{70}{\percent}, whereas Gemini 2.0 Flash with REK (REK+/RAG-) and without REK (REK-/RAG-) showed lower accuracies of \qty{38}{\percent} and \qty{35}{\percent}, respectively (\cref{fig2}). For TNM classification, NotebookLM with REK achieved an accuracy of \qty{80}{\percent}, outperforming Gemini 2.0 Flash with REK (\qty{55}{\percent}) and without REK (\qty{50}{\percent}), with a notable advantage in T and N factors (\cref{fig3}). A similar trend was seen in the classification accuracy of local invasion factors; however, NotebookLM's advantage in resectability classification was not distinct (\cref{fig4}).

    Unlike Gemini 2.0 Flash with and without REK, NotebookLM presented retrieved excerpts from REK as the basis for its classifications, achieving a retrieval accuracy of \qty{92}{\percent} (\cref{fig2}). For example, \cref{fig5} shows the experimental results for Case 98. While Gemini 2.0 Flash with and without REK output only the classification results, which were incorrect, NotebookLM explicitly provided the retrieved REK excerpts as the basis for its correct answers (a subset of the retrieved excerpts is shown in Supplementary file 5\footnote[2]{\label{supplement:results}Supplementary files 1 to 6 can be found in the ancillary files uploaded with this paper on arXiv.}). \cref{fig6} shows the results of Case 48, an example where NotebookLM's retrieval was accurate but its staging was inaccurate. Although sufficient information for correctly determining resectability was retrieved from REK (Supplementary file 6\footref{supplement:results}), the model misclassified resectability based on a description that mistakenly identified the splenic vein (a part of the portal venous system, but not the portal vein (PV) itself) as the PV (\cref{fig6}). There were a few cases (eight in total) in which retrieval from REK was inaccurate. For example, in Case 59, the retrieved excerpts lacked information on ``Resectable: R,'' which was necessary for accurate resectability classification (Supplementary file 2\footref{supplement:results}). Note that the retrieved excerpts for Case 98, Case 48, and Case 59 contain \numlist{678;838;981} words, respectively, accounting for approximately \qtylist{16;19;22}{\percent} of the entire REK (\num{4376} words). These counts, along with those for other cases, can be verified in Supplementary file 2\footref{supplement:results}.

    \begin{figure}
      \begin{tikzpicture}
        \node (input label) {Case 98 CT findings:};
        \node[below=0 of input label.south west, anchor=north west, draw, text width=.5 \textwidth, xshift=12pt] (input) {\vbox{\ttfamily\scriptsize
          An infiltrative pancreatic cancer measuring 41 mm is observed in the tail of the pancreas. Serosal invasion and retropancreatic tissue invasion are noted. Stenosis of the splenic artery is observed, suggesting invasion. Invasion into the left adrenal gland is also noted. No other local invasion factors are identified. Lymph node metastases are observed in three nodes at station 16b. No other metastases are observed.
        }};
        \node[right=6pt of input, align=left, text width=.4 \textwidth] {\vbox{
          Correct answer:\\
          \scriptsize
          Task 1: CH0, DU0, S1, RP1, PV0, A1, PL0, OO1\\
          Task 2: T3\\
          Task 3: N0\\
          Task 4: M1\\
          Task 5: UR
        }};
        \node[below=3pt of input.south west, anchor=north west, xshift=-12pt] (output label) {Output by:};
        \node[below=-3pt of output label.south west, anchor=north west, xshift=12pt] (NotebookLM) {NotebookLM with REK};
        \node[below=0 of NotebookLM.south west, anchor=north west, draw, text width=.8 \textwidth] (NotebookLM output) {\vbox{\ttfamily\scriptsize
          Based on the image findings and the provided information from the sources, here's a breakdown of the tasks:
          \begin{itemize}
            \item Task 1: Diagnose the local invasion factors of pancreatic cancer
            \begin{itemize}
              \item The image findings indicate the following:
              \begin{itemize}
                \item Serosal invasion is present.
                \item Retropancreatic tissue invasion is present.
                \item Invasion into the left adrenal gland is present.
                \item Stenosis of the splenic artery is observed, suggesting arterial system invasion \textcircled{\tiny 1}.
              \end{itemize}
              \item Therefore, based on the definitions in the source \textcircled{\tiny 1}\textcircled{\tiny 2}\textcircled{\tiny 3}.
              \begin{itemize}
                \item CH0, DU0, S1, RP1, PV0, A1, PL0, OO1
              \end{itemize}
            \end{itemize}
            \item Task 2: Determine the T classification of pancreatic cancer
            \begin{itemize}
              \item The tumor is located in the tail of the pancreas and measures 41 mm.
              \item The tumor extends beyond the pancreas but does not involve the CA or SMA \textcircled{\tiny 2}.
              \item According to the JPS eighth edition, T3 is defined as a tumor that extends beyond the pancreas but does not involve the CA or SMA \textcircled{\tiny 2}. T4 is defined as a tumor that involves the CA or SMA.
              \item Therefore, based on the information, the classification is T3.
            \end{itemize}
            \item Task 3: Determine the N classification of pancreatic cancer
            \begin{itemize}
              \item Lymph node metastases are observed in three nodes at station 16b.
              \item Lymph node metastasis outside the regional nodes should be recorded as M1 \textcircled{\tiny 4}.
              \item According to Table 2, for a tumor in the tail of the pancreas, regional lymph nodes include 8a, 9, 10, 11p, 11d, and 18 \textcircled{\tiny 5}.
              \item Lymph node 16b is around the abdominal aorta b \textcircled{\tiny 6}.
              \item Since the metastasis is in station 16b, which is not a regional lymph node for the tail of the pancreas, it should not be considered for N classification \textcircled{\tiny 5}\textcircled{\tiny 7}.
              \item Therefore, the N classification is N0, meaning no regional lymph node metastasis \textcircled{\tiny 8}.
            \end{itemize}
            \item Task 4: Determine the M classification of pancreatic cancer
            \begin{itemize}
              \item Lymph node metastases are observed in three nodes at station 16b \textcircled{\tiny 4}.
              \item Lymph node metastasis outside the regional nodes should be recorded as M1 \textcircled{\tiny 4}.
              \item Therefore, the M classification is M1 \textcircled{\tiny 8}.
            \end{itemize}
            \item Task 5: Determine the resectability classification of pancreatic cancer
            \begin{itemize}
              \item Since distant metastasis, including nonregional lymph node metastasis, is present, the classification is UR-M (tumor with distant metastasis) \textcircled{\tiny 9}.
              \item Therefore, the resectability classification is UR.
            \end{itemize}
          \end{itemize}
        }};
        \node[right=12pt of NotebookLM output] {Correct};
        \node[below=0 of NotebookLM output.south west, anchor=north west] (withREK) {Gemini with REK};
        \node[below=0 of withREK.south west, anchor=north west, draw, text width=.8 \textwidth, align=left] (withREK output) {\vbox{\ttfamily\scriptsize
          Task 1: CH0, DU0, S1, RP1, PV0, A1, PL0, OO1\\
          Task 2: T4\\
          Task 3: N1a\\
          Task 4: M0\\
          Task 5: UR
        }};
        \node[right=12pt of withREK output] {Incorrect};
        \node[below=0 of withREK output.south west, anchor=north west] (withoutREK) {Gemini without REK};
        \node[below=0 of withoutREK.south west, anchor=north west, draw, text width=.8 \textwidth, align=left] (withoutREK output) {\vbox{\ttfamily\scriptsize
          Task 1: CH0, DU0, S1, RP1, PV0, A1, PL0, OO0\\
          Task 2: T4\\
          Task 3: N1b\\
          Task 4: M0\\
          Task 5: UR
        }};
        \node[right=12pt of withoutREK output] {Incorrect};
      \end{tikzpicture}
      \caption{A representative result from the pancreatic cancer staging experiment (Case 98). In this case, both staging and retrieval by NotebookLM were correct, whereas staging by Gemini 2.0 Flash with REK and without REK was incorrect. A subset of the REK excerpts retrieved by NotebookLM is available in Supplementary file 5, while the full set (from \textcircled{\scriptsize 1} to \textcircled{\scriptsize 9}) can be found in Supplementary file 2. \textit{REK}=reliable external knowledge}
      \label{fig5}
    \end{figure}

    \begin{figure}
      \begin{tikzpicture}  
        \node (input label) {Case 48 CT findings:};
        \node[below=0 of input label.south west, anchor=north west, draw, text width=.5 \textwidth, xshift=12pt] (input) {\vbox{\ttfamily\scriptsize
        An infiltrative pancreatic cancer measuring 48 mm is observed in the tail of the pancreas. Serosal invasion and retropancreatic tissue invasion are noted. The splenic vein is obstructed by the tumor, suggesting invasion. No other local invasion factors are identified. No lymph node metastases are present. No other metastases are observed.
        }};
        \node[right=6pt of input, align=left, text width=.4 \textwidth] {\vbox{
          Correct answer:\\
          \scriptsize
          Task 1: CH0, DU0, S1, RP1, PV1, A0, PL0, OO0\\
          Task 2: T3\\
          Task 3: N0\\
          Task 4: M0\\
          Task 5: R
        }};
        \node[below=3pt of input.south west, anchor=north west, xshift=-12pt] (output label) {Output by NotebookLM with REK:};
        \node[below=-6pt of output label.south west, anchor=north west, xshift=12pt] (NotebookLM) {};
        \node[below=0 of NotebookLM.south west, anchor=north west, draw, text width=.8 \textwidth] (NotebookLM output) {\vbox{\ttfamily\scriptsize
        Based on the image findings and the provided information from the sources, here's a breakdown of the tasks:
          \begin{itemize}
            \item Task 1: Diagnose the local invasion factors of pancreatic cancer        
            \begin{itemize}
              \item Based on the image findings, the local invasion factors are:
              \begin{itemize}
                \item Serosal invasion is present \textcircled{\tiny 1}.
                \item Retropancreatic tissue invasion is present \textcircled{\tiny 1}.
                \item Portal venous system (splenic vein) invasion is present \textcircled{\tiny 1}\textcircled{\tiny 2}.
              \end{itemize}
              \item Therefore, the diagnosis is: CH0, DU0, S1, RP1, PV1, A0, PL0, OO0 \textcircled{\tiny 1}.
            \end{itemize}
            \item Task 2: Determine the T classification of pancreatic cancer
            \begin{itemize}
              \item The tumor is located in the tail of the pancreas and measures 48 mm, which is greater than 40 mm but not greater than 60 mm \textcircled{\tiny 3}\textcircled{\tiny 4}.
              \item Serosal and retropancreatic tissue invasion are present \textcircled{\tiny 1}.
              \item Portal venous system invasion is present \textcircled{\tiny 1}\textcircled{\tiny 2}.
              \item According to the JPS eighth edition, T3 is defined as a tumor that extends beyond the pancreas but does not involve the CA or SMA \textcircled{\tiny 1}\textcircled{\tiny 5}. T4 is defined as a tumor that involves the CA or SMA \textcircled{\tiny 1}\textcircled{\tiny 5}.
              \item Given that the tumor extends beyond the pancreas, but does not involve the CA or SMA, the T classification is T3 \textcircled{\tiny 1}.
            \end{itemize}
            \item Task 3: Determine the N classification of pancreatic cancer
            \begin{itemize}
              \item The image findings state that no lymph node metastases are present.
              \item Therefore, the N classification is N0, indicating no regional lymph node metastasis \textcircled{\tiny 6}.
            \end{itemize}
            \item Task 4: Determine the M classification of pancreatic cancer
            \begin{itemize}
              \item The image findings indicate that no distant metastases are observed.
              \item Therefore, the M classification is M0, indicating no distant metastasis \textcircled{\tiny 6}.
            \end{itemize}
    
            \item Task 5: Determine the resectability classification of pancreatic cancer
            \begin{itemize}
              \item The tumor is in the tail of the pancreas.
              \item The tumor invades \ul{the splenic vein (a part of the portal venous system)}, obstructing it \textcircled{\tiny 1}\textcircled{\tiny 2}.
              \item Based on the resectability classifications:
              \begin{itemize}
                \item Resectable (R) indicates no tumor contact with the SMV or PV, or less than 180 degrees contact/invasion without occlusion \textcircled{\tiny 7}.
                \item \ul{Borderline Resectable (BR-PV)} indicates tumor contact/invasion of the SMV/PV of 180 degrees or more, or occlusion of the SMV/PV, not exceeding the inferior border of the duodenum \textcircled{\tiny 7}.
                \item Unresectable (UR-LA) indicates tumor contact/invasion of the SMV/PV of 180 degrees or more, or occlusion of the SMV/PV extending beyond the inferior border of the duodenum \textcircled{\tiny 8}.
              \end{itemize}
              \item \ul{Since the splenic vein is obstructed, the tumor is considered at least borderline resectable} \textcircled{\tiny 7}. However, since the image findings do not clarify whether the occlusion extends beyond the inferior border of the duodenum, it is difficult to distinguish between BR and UR. Therefore, the classification is BR.
            \end{itemize}
          \end{itemize}
        }};
        \node[below=0 of NotebookLM output] {\textcolor{red}{The LLM misinterpreted the splenic vein as the portal vein (PV).}};
        \node[right=12pt of NotebookLM output, align=center] {
          Incorrect\\
        (Task 5)};
      \end{tikzpicture}
      \caption{A representative result from the pancreatic cancer staging experiment (Case 48). In this case, although NotebookLM appropriately retrieved REK excerpts, the LLM misinterpreted the information, leading to an incorrect staging. A subset of the REK excerpts retrieved by NotebookLM is available in Supplementary file 5, while the full set (from \textcircled{\scriptsize 1} to \textcircled{\scriptsize 8}) can be found in Supplementary file 2. \textit{REK}=reliable external knowledge, \textit{LLM}=large language model}
      \label{fig6}
    \end{figure}

  \section{Discussion}
    Several previous studies have tried pancreatic cancer staging from radiology reports using LLMs. Bhayana R. et al. evaluated GPT-3.5 and GPT-4 in generating synoptic radiology reports and assessing tumor resectability for pancreatic ductal adenocarcinoma. Their study found that GPT-4 produced near-perfect synoptic reports, and chain-of-thought prompting improved its accuracy in classifying resectability. However, the model still made some clinically significant misinterpretations that could affect decision-making, so the authors argue that LLM-based applications should be used only as aids under supervised settings \cite{Bhayana1}. Suzuki K. et al. assessed GPT-4's capability in TNM classification of pancreatic cancer using Japanese radiology reports, and the authors concluded that its performance did not meet clinical standards \cite{Suzuki}. Similarly, in our experimental settings, pancreatic cancer staging by Gemini 2.0 Flash exhibited poor performance. In contrast, providing the pancreatic cancer staging guidelines in Japan as REK in the prompt field led to a slight improvement in accuracy. Moreover, a clear improvement was observed in NotebookLM, where Gemini 2.0 Flash was enhanced with RAG technology (\cref{fig2,fig3,fig4}). The superiority of NotebookLM in cancer staging was also demonstrated in our previous study on lung cancer \cite{Tozuka}. Taken together, these findings suggest that RAG technology may enhance the accuracy of cancer staging by LLMs.

    Although NotebookLM exhibited superior performance in cancer staging, classification errors persisted even when relevant excerpts from REK were appropriately retrieved, as the LLM misinterpreted the information (\cref{fig6}, Supplementary file 6\footnote[2]{\label{supplement:discussion}Supplementary files 1 to 6 can be found in the ancillary files uploaded with this paper on arXiv.}). Misinterpretations (or hallucinations) remain a serious concern in the field of LLM research, and no complete solution has been found \cite{Huang}. Therefore, even a highly accurate LLM poses risks if used in medical practice without physician oversight, reinforcing the need to limit its role to a supplementary tool. For guideline-based tasks like cancer staging, LLM-generated classifications without reliable supporting evidence are unlikely to reduce physicians' workload, as they would still need to consult the guidelines just as they would without LLM assistance. However, this study demonstrated that the RAG-LLM, NotebookLM, can explicitly provide retrieved REK excerpts as evidence for staging (see \cref{fig5,fig6}, as well as Supplementary files 5 and 6\footref{supplement:discussion} for examples) with relatively high accuracy (\cref{fig2}), potentially allowing physicians to verify facts more efficiently. Future studies should evaluate whether the partial provision of REK through RAG technology effectively alleviates physicians' burden. 

    As discussed above, RAG-LLMs have the potential to assist physicians with diagnosis and classification; however, data management issues remain a concern for practical medical applications. Submitting patient information, such as radiology reports, to internet-based LLMs like Gemini (Google) or GPT (OpenAI) is generally discouraged due to information security concerns. In light of this, open-source LLMs that can be downloaded and run locally are increasingly being recommended in the field of radiology \cite{Savage,Nowak}. Therefore, in the future, it will be necessary to pursue the clinical application of offline or on-premises RAG-LLMs instead of NotebookLM.

    The remaining limitations and future perspectives of this study are as follows. First, the extent to which RAG enhances LLM performance may depend on the LLM's capability as well as the length and complexity of the REK. If the LLM can fully comprehend and process the entire REK, directly inputting it into the prompt field may seem sufficient, potentially leading to the assumption that partial information extraction, as performed in RAG, is unnecessary. Nevertheless, if the REK is too extensive for physicians, extracting relevant information using RAG techniques may still be beneficial in assisting physicians with staging. Second, the retrieval accuracy we defined is affected by the length of the REK. In our experiment, the total word count of the retrieved excerpts for each case was roughly \qty{20}{\percent} of the REK (see Supplementary file 2\footref{supplement:discussion} for raw data). However, if the excerpts encompassed the entire REK, the retrieval accuracy would inevitably reach \qty{100}{\percent}, making the result trivial (which did not occur in this experiment). Therefore, assessing retrieval accuracy based solely on its numerical value is not appropriate. Third, unlike in actual clinical settings, we evaluated the LLM's staging accuracy using fictional cancer CT findings and Japan's pancreatic cancer staging guidelines in English rather than Japanese. To rigorously assess its applicability in real clinical practice, validation with actual clinical data is necessary. Fourth, while our previous study \cite{Tozuka} and this research have evaluated the utility of a RAG-LLM (NotebookLM) in cancer staging, future studies should also explore its applicability to other clinical tasks, such as identifying differential diagnoses based on imaging findings.

  \section{Conclusion}
    NotebookLM, a RAG-LLM, demonstrated superior accuracy in a pancreatic cancer staging experiment compared to Gemini 2.0 Flash. Since Gemini 2.0 Flash is also the underlying LLM used in NotebookLM, this result suggests that RAG technology has the potential to enhance the staging accuracy of LLMs. Additionally, NotebookLM effectively retrieved relevant excerpts from the given REK, allowing physicians to assess the reliability of its responses and highlighting the potential of RAG-LLMs in supporting clinical diagnosis and classification. Due to information security concerns, the internet-based RAG-LLMs, including NotebookLM, are unlikely to be suitable for medical applications, underscoring the need for locally operating RAG-LLMs in the future.

  \section{Declarations}
    \subsection*{Author contributions}
      \begin{itemize}[label=\textbullet, leftmargin=*]
        \item Conceptualization: Hisashi Johno
        \item Methodology: Hisashi Johno, Ryota Tozuka, Hiroaki Watanabe
        \item Investigation: Junichi Sato, Yuki Johno, Akitomo Amakawa, Hisashi Johno, Ryota Tozuka, Atsushi Komaba
        \item Validation: Akitomo Amakawa, Yuki Johno, Hiroaki Watanabe, Hiroki Watanabe, Chihiro Goto, Hiroyuki Morisaka
        \item Formal analysis: Yuki Johno, Akitomo Amakawa
        \item Visualization: Yuki Johno, Hisashi Johno
        \item Writing---original draft: Yuki Johno, Hisashi Johno
        \item Writing---review \& editing: Hisashi Johno, Atsushi Komaba, Akitomo Amakawa, Hiroyuki Morisaka, Ryota Tozuka, Kazunori Nakamoto, Hiroaki Watanabe, Junichi Sato
        \item Supervision: Hisashi Johno, Kazunori Nakamoto
        \item Project administration: Hisashi Johno
        \item Resources: Hisashi Johno, Atsushi Komaba, Kazunori Nakamoto, Hiroshi Onishi
        \item Funding acquisition: Hisashi Johno, Kazunori Nakamoto
      \end{itemize}
    \subsection*{Acknowledgements}
      We sincerely appreciate Dr. Tsutomu Fujii (MD, PhD, FACS), Professor and Chairman of the Department of Surgery and Science, Faculty of Medicine, Academic Assembly, University of Toyama, for his valuable guidance on the contents of the Japanese Classification of Pancreatic Carcinoma, Eighth Edition by the Japan Pancreas Society.
    \subsection*{Funding}
      This study was partially supported by JSPS KAKENHI Grant Numbers JP21K15762 and JP24K06686.
    \subsection*{Competing interests}
      There are no competing interests with regard to this manuscript.
    \subsection*{Ethics approval}
      Since the study used only fictional patient data, ethical approval was not required.
    \subsection*{Informed consent}
      Not applicable.
    \subsection*{Data availability statement}
      Most of the data supporting the findings of this study are included in the article and its Supplementary files. Further details and additional data can be obtained from the corresponding author upon reasonable request.

  \bibliographystyle{unsrt}
  \bibliography{manuscript}

\begin{thebibliography}{10}

\bibitem{Keshavarz}
Pedram Keshavarz, Sara Bagherieh, Seyed~Ali Nabipoorashrafi, Hamid Chalian,
  Amir~Ali Rahsepar, Grace Hyun~J. Kim, Cameron Hassani, Steven~S. Raman, and
  Arash Bedayat.
\newblock {ChatGPT} in radiology: A systematic review of performance, pitfalls,
  and future perspectives.
\newblock {\em Diagnostic and Interventional Imaging}, 105(7--8):251--265,
  2024.

\bibitem{Lewis}
Patrick Lewis, Ethan Perez, Aleksandra Piktus, Fabio Petroni, Vladimir
  Karpukhin, Naman Goyal, Heinrich K{\"u}ttler, Mike Lewis, Wen tau Yih, Tim
  Rockt{\"a}schel, Sebastian Riedel, and Douwe Kiela.
\newblock Retrieval-augmented generation for knowledge-intensive {NLP} tasks.
\newblock In {\em NIPS '20: Proceedings of the 34th International Conference on
  Neural Information Processing Systems}, pages 9459--9474, 2020.

\bibitem{Shuster}
Kurt Shuster, Spencer Poff, Moya Chen, Douwe Kiela, and Jason Weston.
\newblock Retrieval augmentation reduces hallucination in conversation.
\newblock In {\em Findings of the Association for Computational Linguistics:
  EMNLP 2021}, pages 3784--3803, 2021.

\bibitem{Bhayana2}
Rajesh Bhayana, Aly Fawzy, Yangqing Deng, Robert~R. Bleakney, and Satheesh
  Krishna.
\newblock Retrieval-augmented generation for large language models in
  radiology: Another leap forward in board examination performance.
\newblock {\em Radiology}, 313(1):e241489, 2024.

\bibitem{Rau}
Alexander Rau, Stephan Rau, Daniela Z{\"o}ller, Anna Fink, Hien Tran, Caroline
  Wilpert, Johanna Nattenm{\"u}ller, Jakob Neubauer, Fabian Bamberg, Marco
  Reisert, and Maximilian~F. Russe.
\newblock A context-based chatbot surpasses radiologists and generic {ChatGPT}
  in following the {ACR} appropriateness guidelines.
\newblock {\em Radiology}, 308(1):e230970, 2023.

\bibitem{Tozuka}
Ryota Tozuka, Hisashi Johno, Akitomo Amakawa, Junichi Sato, Mizuki Muto,
  Shoichiro Seki, Atsushi Komaba, and Hiroshi Onishi.
\newblock Application of {NotebookLM}, a large language model with
  retrieval-augmented generation, for lung cancer staging.
\newblock {\em Japanese Journal of Radiology}, 2024.
\newblock Online ahead of print.

\bibitem{Ishida}
Masaharu Ishida, Tsutomu Fujii, Masashi Kishiwada, Kazuto Shibuya, Sohei Satoi,
  Makoto Ueno, Kohei Nakata, Shigetsugu Takano, Katsunori Uchida, Nobuyuki
  Ohike, Yohei Masugi, Toru Furukawa, Kenichi Hirabayashi, Noriyoshi Fukushima,
  Shuang-Qin Yi, Hiroyuki Isayama, Takao Itoi, Takao Ohtsuka, Takuji Okusaka,
  Dai Inoue, Hirohisa Kitagawa, Kyoichi Takaori, Masaji Tani, Yuichi Nagakawa,
  Hideyuki Yoshitomi, Michiaki Unno, and Yoshifumi Takeyama.
\newblock Japanese classification of pancreatic carcinoma by the {Japan}
  pancreas society: Eighth edition.
\newblock {\em Journal of Hepato-Biliary-Pancreatic Sciences}, 31(11):755--768,
  2024.

\bibitem{Bhayana1}
Rajesh Bhayana, Bipin Nanda, Taher Dehkharghanian, Yangqing Deng, Nishaant
  Bhambra, Gavin Elias, Daksh Datta, Avinash Kambadakone, Chaya~G. Shwaartz,
  Carol-Anne Moulton, David Henault, Steven Gallinger, and Satheesh Krishna.
\newblock Large language models for automated synoptic reports and
  resectability categorization in pancreatic cancer.
\newblock {\em Radiology}, 311(3):e233117, 2024.

\bibitem{Suzuki}
Kazufumi Suzuki, Hiroki Yamada, Hiroshi Yamazaki, Goro Honda, and Shuji Sakai.
\newblock Preliminary assessment of {TNM} classification performance for
  pancreatic cancer in {Japanese} radiology reports using {GPT}-4.
\newblock {\em Japanese Journal of Radiology}, 43(1):51--55, 2025.

\bibitem{Huang}
Lei Huang, Weijiang Yu, Weitao Ma, Weihong Zhong, Zhangyin Feng, Haotian Wang,
  Qianglong Chen, Weihua Peng, Xiaocheng Feng, Bing Qin, and Ting Liu.
\newblock A survey on hallucination in large language models: Principles,
  taxonomy, challenges, and open questions.
\newblock {\em ACM Transactions on Information Systems}, 43(2):No. 42, 2025.

\bibitem{Savage}
Cody~H. Savage, Adway Kanhere, Vishwa Parekh, Curtis~P. Langlotz, Anupam Joshi,
  Heng Huang, and Florence~X. Doo.
\newblock Open-source large language models in radiology: A review and tutorial
  for practical research and clinical deployment.
\newblock {\em Radiology}, 314(1):e241073, 2025.

\bibitem{Nowak}
Sebastian Nowak, Benjamin Wulff, Yannik~C. Layer, Maike Theis, Alexander Isaak,
  Babak Salam, Wolfgang Block, Daniel Kuetting, Claus~C. Pieper, Julian~A.
  Luetkens, Ulrike Attenberger, and Alois~M. Sprinkart.
\newblock Privacy-ensuring open-weights large language models are competitive
  with closed-weights {GPT}-4o in extracting chest radiography findings from
  free-text reports.
\newblock {\em Radiology}, 314(1):e240895, 2025.

\end{thebibliography}
\end{document}